\documentclass[runningheads]{llncs}
\usepackage{graphicx}
\usepackage{siunitx}
\usepackage[pdftex,
    colorlinks={true},
    linkcolor={blue},
    citecolor={black},
    bookmarksopenlevel=1,
]{hyperref}	

\usepackage[space]{cite}    

\def\PFlong{{\em Ranitomeya imitator}}
\def\PF{{\em R. imitator~}}

\usepackage{gensymb}

\begin{document}
\title{Feed Me: Robotic Infiltration\\ of Poison Frog Families}
%
%
\author{Tony G. Chen\inst{1} \and
Billie C. Goolsby\inst{2} \and
Guadalupe Bernal\inst{1} \and \\
Lauren A. O'Connell\inst{2} \and
Mark R. Cutkosky\inst{1}}
\authorrunning{T. G. Chen et al.}
%
\institute{(1) Dept.~of Mechanical Engineering, (2) Dept.~of Biology, Stanford University, Stanford CA 94305, USA
\email{agchen@stanford.edu}, 
\url{http://bdml.stanford.edu}}
\maketitle              
\begin{abstract}
We present the design and operation of tadpole-mimetic robots prepared for a study of the parenting behaviors of poison frogs, which pair bond and raise their offspring. The mission of these robots is to convince poison frog parents that they are tadpoles, which need to be fed. Tadpoles indicate this need, at least in part, by wriggling with a characteristic frequency and amplitude. While the study is in progress, preliminary indications are that the TadBots have passed their test, at least for father frogs. We discuss the design and operational requirements for producing convincing TadBots and provide some details of the study design and plans for future work.
\end{abstract}
\section{Introduction}
\label{sec:intro}
\begin{figure}
\centering
\includegraphics[width=1\textwidth]{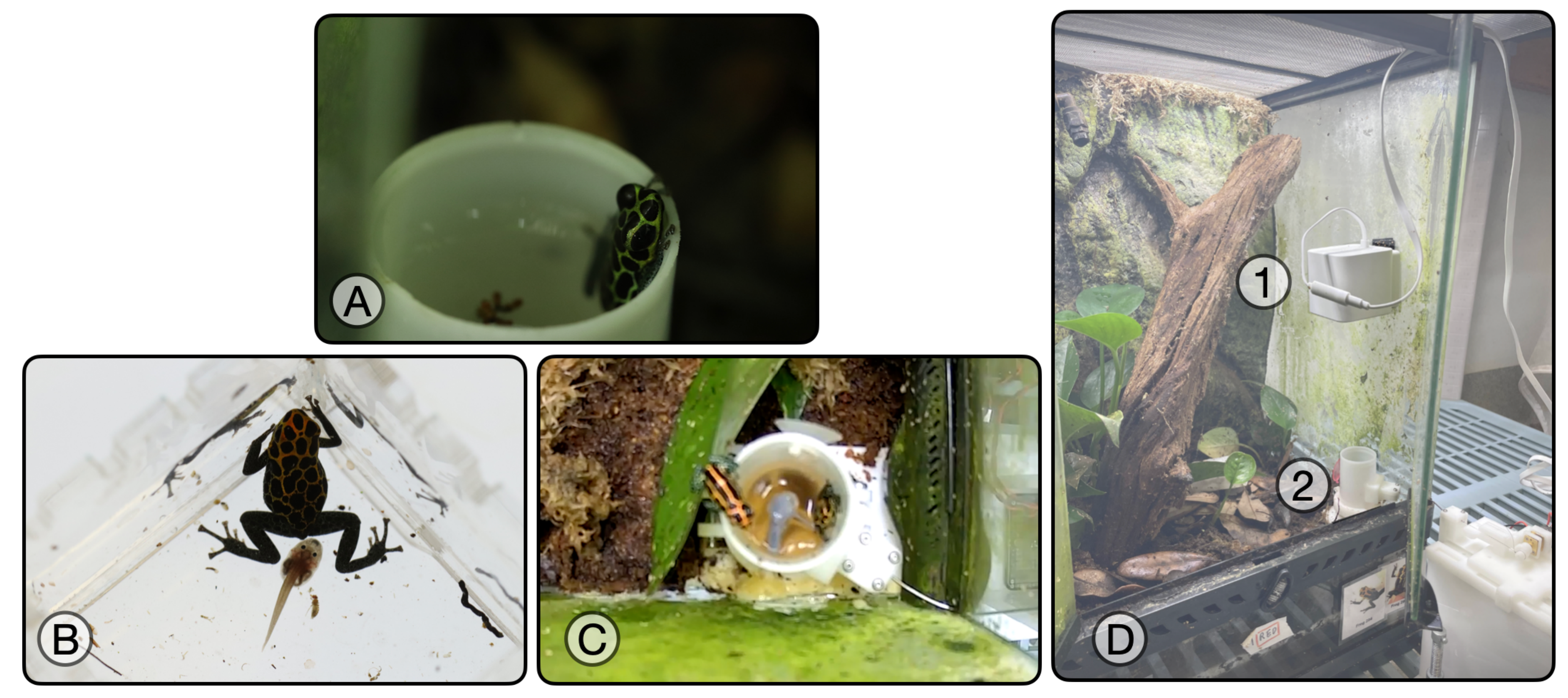}
\caption{Infiltration of Poison Frog Families with TadBots. (A) Frogs enter the pools of water with their heads facing away from the tadpole. (B) Young tadpoles, approximately 3/4 the length of a frog, approach the vent of the frog to beg. (C) Parents attempt to coordinate TadBot care. (D) Typical experiment chamber with camera 
 \raisebox{.5pt}{\textcircled{\raisebox{-.9pt} {1}}} positioned above a TadBot canister \raisebox{.5pt} {\textcircled{\raisebox{-.9pt} {2}}}; the canister is identical to those used for biological offspring. Photos A-B courtesy of Daniel Shaykevich.
}\label{fig1}
\vspace{-10mm}
\end{figure}

Complex behavioral interactions govern animal social life, especially in relation to parental care and coordination critical for the survival of a species. The mimetic poison frog, \PFlong, a monogamous poison frog native to the north-central region of eastern Peru, is biparental, meaning that both mothers and fathers must work together as a team for their offspring to have the greatest chances of survival \cite{brown2011taxonomic,brown2008divergence,brown2010key}. \PF fathers transport their tadpoles piggy-back style into pools of water situated in bromeliad plant cavities, which they visit and guard at least daily \cite{brown2011taxonomic,brown2008divergence,brown2010key}. The father deposits one tadpole per pool 
because the tadpoles are cannibalistic to their sibling conspecifics as a consequence of their low-resource environments \cite{brown2008divergence}. 
At any time, \PF parents normally care for one to three tadpoles \cite{yoshioka2016evidence}. When the father observes that a tadpole needs to be fed, he calls for his partner to provision one to two unfertilized egg meals \cite{moss2023evolution,yoshioka2016evidence}. The tadpoles signal that they are in need of nutritional resources by intensely wriggling, which both parents can observe \cite{yoshioka2016evidence}. When a frog makes contact with the pond, the tadpole also vibrates against the frog's abdomen to elicit care. Rather than using kin recognition, poison frogs use spatial memory of the pool sites to determine which tadpoles to provide with care \cite{ringler2017adopt,ringler2016sex,stynoski2009discrimination}.
We exploit this characteristic to add robotic tadpole infiltrators into poison frog families, to study parenting and explore which tadpole signals are relevant to care (Fig.~\ref{fig1}).

In other work, model frogs, robotic frogs, and even electrodynamic shakers have been used across multiple species to test social decision-making, including treefrogs and tungara frogs  \cite{brown2008divergence,jungfer1996reproduction,coss2022can,klein2012robots,taylor2008faux,caldwell2010vibrational}. In the present case, we are interested in producing tadpole-mimetic robots that can influence parental decision-making. In this context, the test for a robot is whether it can convince 
\PF parents that it is a tadpole that needs to be guarded and fed.


\subsection{Characteristics of begging in \PFlong}

Begging as a form of parent-offspring communication has independently evolved multiple times across the vertebrate lineages and also within the Amphibia class \cite{dugas2017tadpole,jungfer1996reproduction,kam2002female,yoshioka2016evidence,wright2007evolution}. However, research on what begging actually signals in poison frogs has reached conflicting conclusions. In the strawberry poison frog, \emph{Oophaga pumilio}, which is a female uniparental system, only tadpoles of greater fitness beg, meaning that begging is a signal of quality rather than need \cite{dugas2017tadpole}. In \PF\!\!, which are biparental and spread the parental burden of care, smaller and more nutritionally needy tadpoles beg more, suggesting, that begging in \PF is a signal of need \cite{yoshioka2016evidence}. These conflicting signals of quality versus need reflect a theoretical schism concerning the function of begging in parental decision-making. In either case, what has not been demonstrated is whether begging \emph{intensity} acts as a signal that influences parental care.

To investigate whether begging intensity is a signal that influences parental effort, we need a tractable method of modifying single or multiple features of the begging signal. \PF poison frogs are an ideal model to investigate what information begging signals contain, as poison frogs do not recognize individual offspring but remember the spatial locations of their nurseries \cite{stynoski2009discrimination}. Therefore, we hypothesized that it would be possible to cross-foster biological offspring with robotic tadpole imposters. 

In order to maximize our chances with a robotic infiltrator, we first considered the sensory modalities that poison frogs likely employ when interacting with their offspring: olfactory, visual, and tactile. Nursery water for tadpoles is often dirty, as it contains detritus, algae, and dirt -- all of which are desirable to poison frogs for hosting offspring in well-resourced environments with stagnant water. Studies suggest that poison frogs likely prioritize olfactory cues in decision-making about which pools of water to populate \cite{serrano2021tadpole}. Other studies across Anurans have shown that touch, especially vibrational processing, is critical for life-or-death decision-making \cite{dugas2017tadpole,hill2022biotremology}. In red-eyed treefrogs, vestibular mechanoreception dictates the escape-hatch response of embryos \cite{jung2019red}. Mechanoreception presents an evolutionary response to detecting predators like snakes, that may eat embryos growing on rainforest canopy leaves \cite{jung2019red}. Early experimental studies showed that vibration
-- not olfaction or vision -- was the necessary cue to stimulate egg feeding behavior \cite{jungfer1996reproduction}. In summary, these findings illustrate that vibration and touch are an important language for frogs with the capacity to have different meanings in different social contexts.


Begging in \PF is a dazzling visual and tactile display. During six minutes of exposure to mothers, tadpoles can beg for 1-4 minutes intensely wriggling and vibrating their bodies against a parent entering the nursery \cite{yoshioka2016evidence}. For comparison, studies have shown in other species of poison frog tadpoles that the mean duration of a begging bout is 12-15 seconds\cite{dugas2017tadpole}. 

\subsection{Design Requirements}
\label{design-reqs}
\vspace{-3mm}

As noted above, olfactory, visual, and haptic (vibrational and tactile) cues evidently play a role in \PF parenting. To match olfactory signals, the robot should function in tadpole-conditioned water, achieved when a tadpole has lived in the water for at least \SI{24}{\hour}, supplemented with detritus such as waste from frogs and dead flies, which are common to tadpole nurseries. 
For scale, the neutrally buoyant body should be roughly 75\%--100\% of the length of a parent (adult size: 16.0 - 17.5mm, \cite{brown2011taxonomic}) to proportionally mimic a tadpole between the Gosner stages 30-40 \cite{gosner1960simplified}.
To encapsulate the body we require a soft material to match the feel of a tadpole's skin which contains its viscera. It is also desirable to match the tadpole's color, as poison frogs appear to rely on contrast for visual detection. Finally, we want to mimic the stereotypical begging motion
in which the tail undulates with respect to the head, which also vibrates side-to-side. We desire to match the frequencies, amplitudes, and durations recorded in previous observations of tadpoles \cite{dugas2017tadpole} and our own observations \cite{coss2022can}. These parameters are summarized in Table \ref{parameters} and govern the mechanism design in Section \ref{mechanism}. 

\begin{table}[]
    \centering
    \caption{Required Parameters for TadBot Design}
    \begin{tabular}{||c | c || c | c||}
        \hline
        Requirement & Range & Requirement & Range \\ [0.5ex]
        \hline\hline
        overall length & $\leq$20 mm & mass & 0.15-0.3 g \\
        \hline
        head major diameter & $\approx$8 mm & minor diameter & $\approx$6 mm \\
        \hline
        oscillation freqs. & 5-25 Hz & amplitude & $\approx$5mm \\
        \hline
        skin & dark gray/brown & hardness & $\approx$Shore A 00-20 \\ [1ex]
        \hline
    \end{tabular}
    \vspace{6pt}
    \label{parameters}
\end{table}
\vspace{-4mm}


\section{Methods}
\subsection{Mechanism Design}
\label{mechanism}
\begin{figure}
\centering
\includegraphics[width=0.9\textwidth]{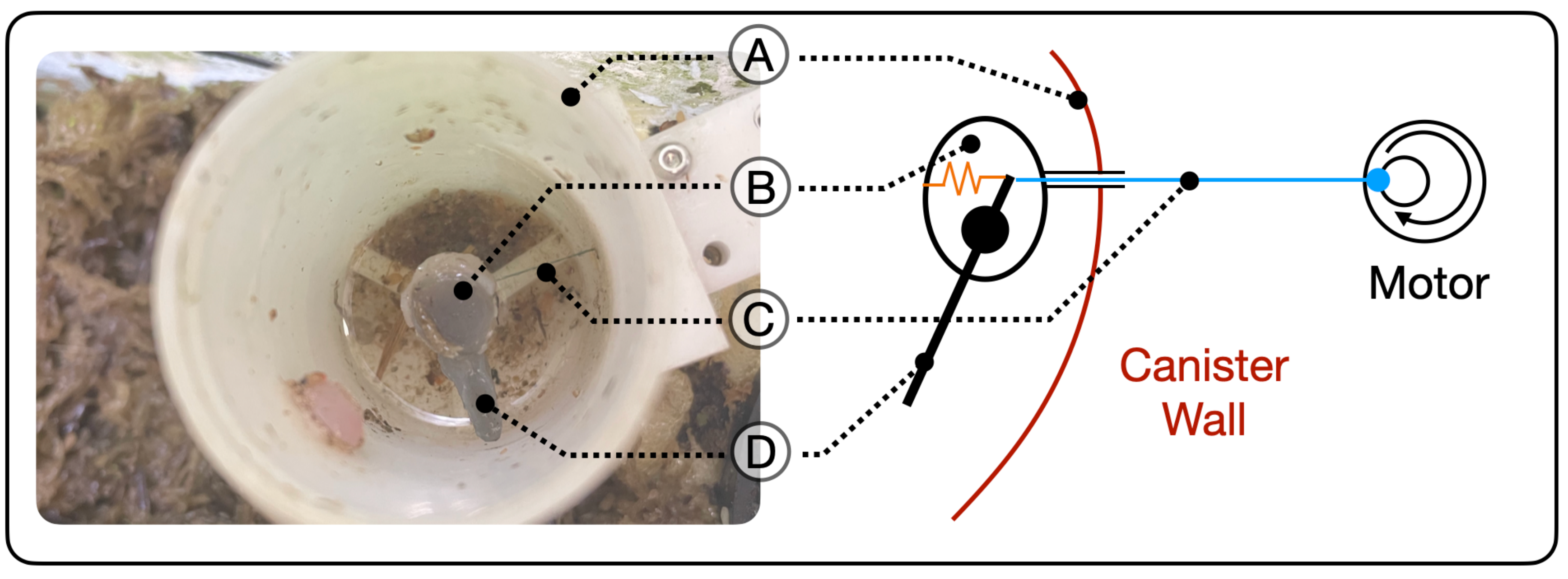}
\caption
{Tadbot resides inside a plastic canister (A). The body (B) encloses a tail lever (D) that rotates about a pivot under the action of a motor-driven tendon (C) and restoring elastic band.}
\label{fig:mech}
\vspace{-6mm}
\end{figure}

\begin{figure}
\centering
\includegraphics[width=0.9\textwidth]{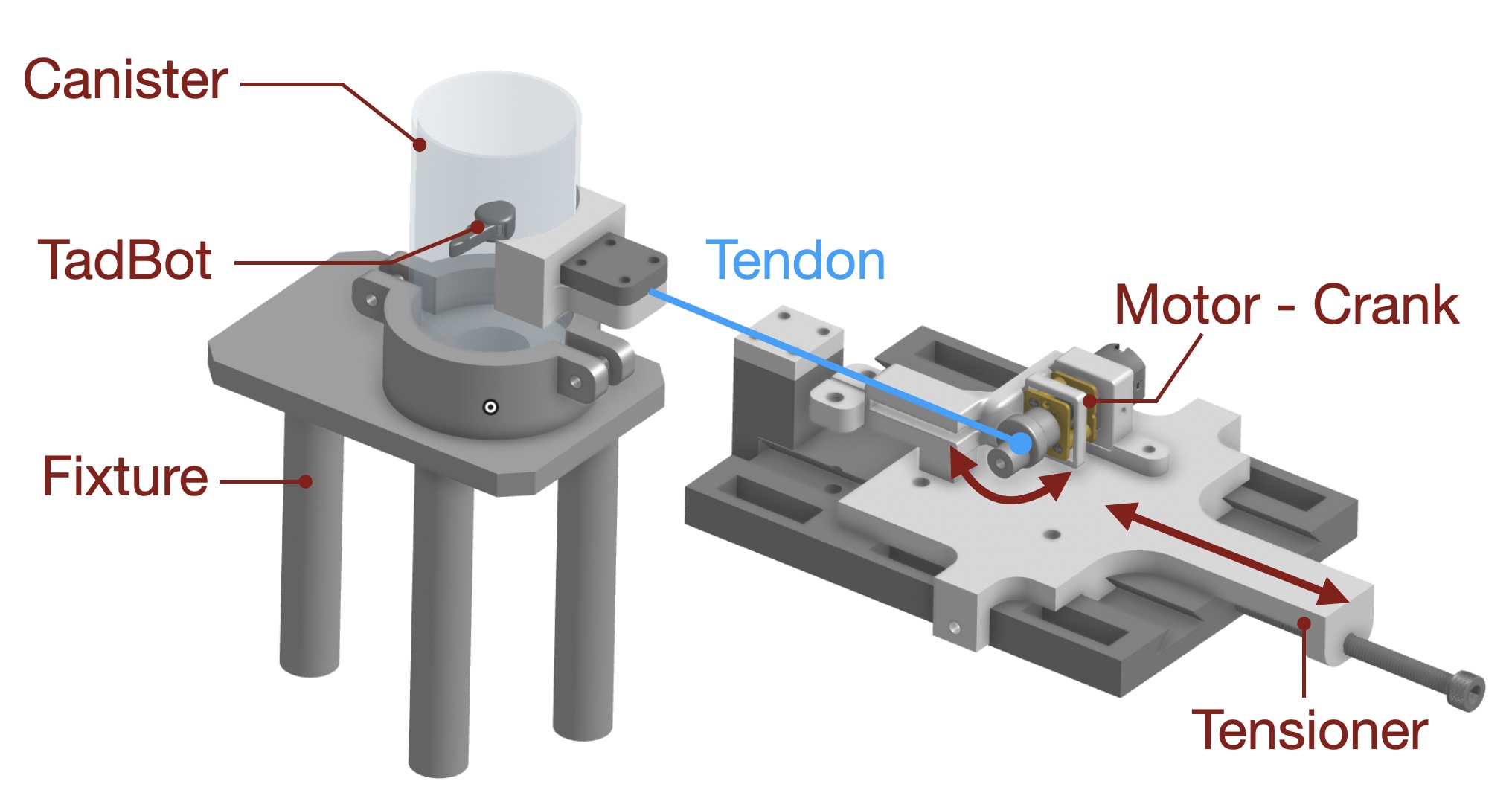}
\caption
{CAD rendering of TadBot system. TadBot is suspended inside a canister, mounted on a platform in the terrarium where the parents live. The actuation assembly is mounted remotely and consists of a motor-crank mechanism and tensioner.}\label{mech}
\vspace{-6mm}
\end{figure}

To meet the design requirements, we have designed and built TadBots that mimic the appearance and begging dynamics of \PF tadpoles. The body of the TadBot has four major components (Fig.~\ref{fig:mech}). To keep the body small, and isolate any noise and high-frequency motor vibrations from the nursery canister, the TadBot is driven remotely by a motor and crank mechanism that connects to the body using a \SI{30}{\centi\meter} long tendon running through a soft plastic sleeve (Fig.~\ref{mech}C).
The tendon acts upon a lever inside the TadBot body that rotates about a dowel pin as a pivot. An elastic band maintains tension and restores the lever position as the tendon relaxes.

The TadBot body is suspended inside a water-filled plastic canister, the usual habitat of \PF tadpoles in the laboratory. This is achieved by having the soft plastic sleeve glued to the TadBot body at one end and to the canister wall at the other end. The tendon terminates at the motor-crank and tensioner mechanism (Fig.~\ref{mech}). TadBots have a mass of \SI{0.3}{\gram} and outer dimensions of \SI{20}{\milli\meter} by \SI{8}{\milli\meter} by \SI{5}{\milli\meter} (LxWxD).

For adjustment, the motor-crank assembly is mounted on a platform that slides on a fixed stand, and the relative position of the two can be adjusted by turning a tensioning screw to ensure that (i) the motor-crank assembly is providing enough range of motion to the oscillating lever and (ii) the tendon tension does not exceed the buckling strength of the plastic sleeve.

\subsection{Body and Skin}

\begin{figure}
\centering
\includegraphics[width=0.9\textwidth]{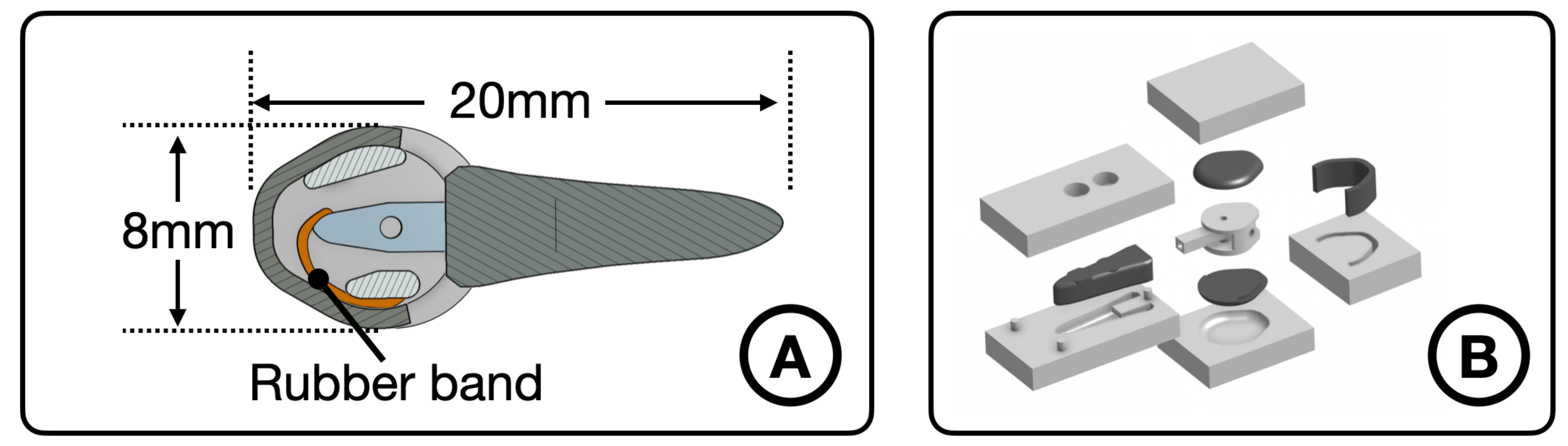}
\caption
{TadBot fabrication: (A) Moving components (lever and rubber band) are contained in a SmoothOn Eco-Flex 00-20 skin. (B) The assembly is cast from multiple 3D prointed molds and glued using cyanoacrylate adhesive.}
\label{manufacture}
\vspace{-6mm}
\end{figure}

A soft silicone skin approximates the texture and feel of tadpole skin when it comes into contact with a parent. A black pigment is mixed into Ecoflex 00-20 to match the dark gray skin tone of a tadpole. The skin is \SI{1}{\milli\meter} thick and is made in pieces, shown in Fig.~\ref{manufacture}B. The top and bottom pieces are identical. The side piece is molded to provide the desired profile and enclose the moving parts. The tail is made from a two-part mold so that it can slip onto the end of the oscillating lever. The silicone pieces are glued together with cyanoacrylate adhesive (it is not necessary to form a watertight seal).

\subsection{Actuation Dynamics}

\begin{figure}
\centering
\includegraphics[width=0.9\textwidth]{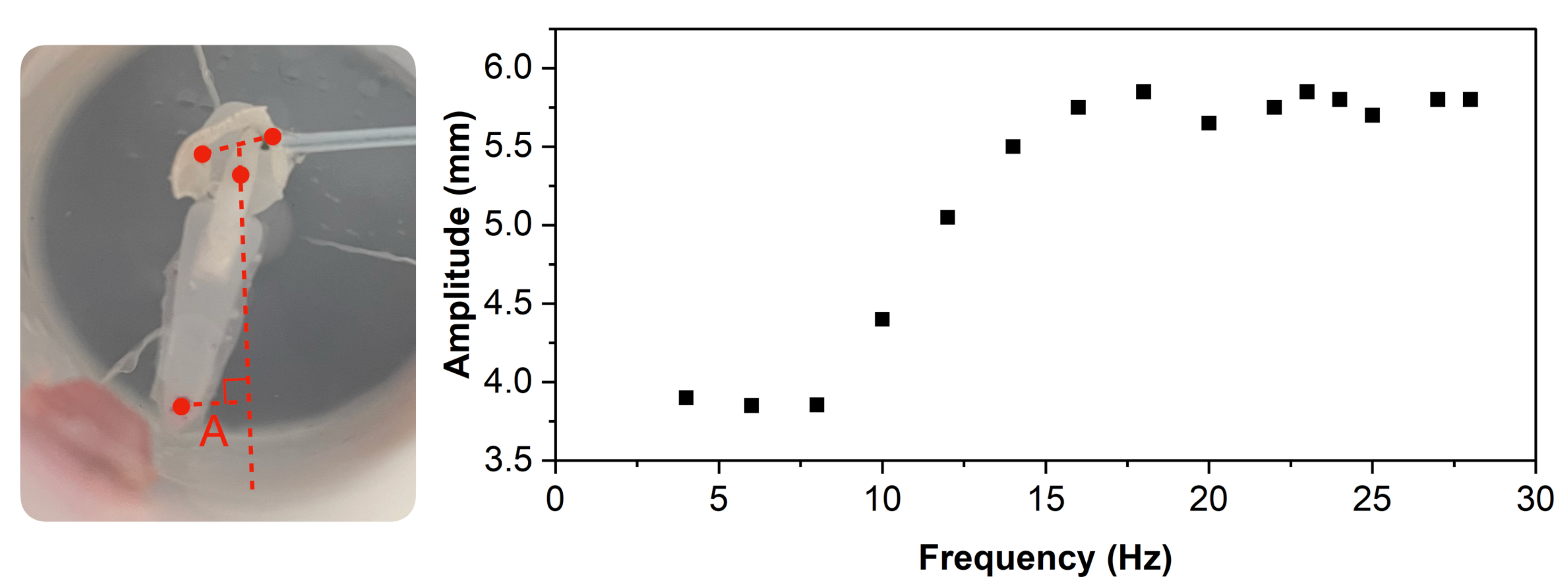}
\caption{Relationship between wiggling frequency and amplitude (A) of the tail displacement, measured with a camera at 240\,fps tracking markers on the head and tail.}\label{fig3}
\vspace{-8pt}
\end{figure}

A frequency and amplitude characterization was conducted to ensure that TadBot's tail achieves the desired wiggling displacement at the desired operating frequencies (Table~\ref{parameters}). Two markers were drawn on the head of a TadBot spaced \SI{4.2}{\milli\meter} apart, with a \SI{15}{\degree} offset from the transverse plane (Fig.~\ref{fig3}). Then a line is drawn from the bisection point between these two points and the pivot point of the tail to establish the median line along the sagittal plane. An additional marker is placed at the tip of the tail, and the amplitude is measured between this marker and the median line. The results are plotted in Fig.~\ref{fig3}. 

At frequencies below 8\,Hz, consistent with gentle, non-begging swimming, the tail appears relatively free of inertial effects and undergoes a low amplitude oscillation. As the frequency increases above \SI{10}{\hertz}, there is some additional displacement due to the inertia of the silicone tail. The behavior, however, is not noticeably resonant, and the amplitude plateaus between \SI{15}-\SI{28}{\hertz}, which covers the upper limit of observed begging frequencies in tadpoles (Table~\ref{parameters}). 

\subsection{Experiment Setup and Procedure}
Multiple aquarium tanks are set up for the pair-bonded frog parents (Fig~\ref{fig1}). Tadpoles inside the habitat are swapped out with TadBots after successful feeding behaviors from the parents are observed. 
To normalize care efforts across pairs, the number of tadpoles was controlled by limiting the number of nurseries available for parents to deposit. Any extra tadpoles deposited were subsequently removed from the tank. The experiment setup is shown in Fig.~\ref{fig1}D.

All \PF 
used in the laboratory study were captive-bred in our poison frog colony or purchased from Ruffing’s Ranitomeya (Tiffon, Ohio, USA). One adult male and female are housed together in a 45.72 x 30.48 x 30.48 cm terrarium (Exoterra, Rolf C. Hagen USA, Mansfield, MA) containing sphagnum moss substrate, driftwood, live Pothos plants, horizontally mounted film canisters as egg deposition sites, and additional film canisters filled with water, treated with reverse osmosis (R/O Rx, Josh’s Frogs, Owosso, MI) for tadpole deposition.
Terraria were automatically misted ten times daily for 20 seconds each, and frogs were fed live \emph{Drosophila melanogaster} flies dusted with vitamin powder, thrice weekly. The tanks were also supplemented with \emph{Folsomia candida} and \emph{Trichorhina tomentosa}. The observation housing was set on a 12:12 light cycle from 07:00 to 19:00\,hrs. The average temperature and humidity were recorded for each day of observation, usually around \SI{25}{\degree} and \SI{95}{\percent} humidity within the tank. The experiment is approved under Stanford APLAC protocol 34242. 

\begin{figure}[h!]%
\centering
\includegraphics[width=0.8\textwidth]{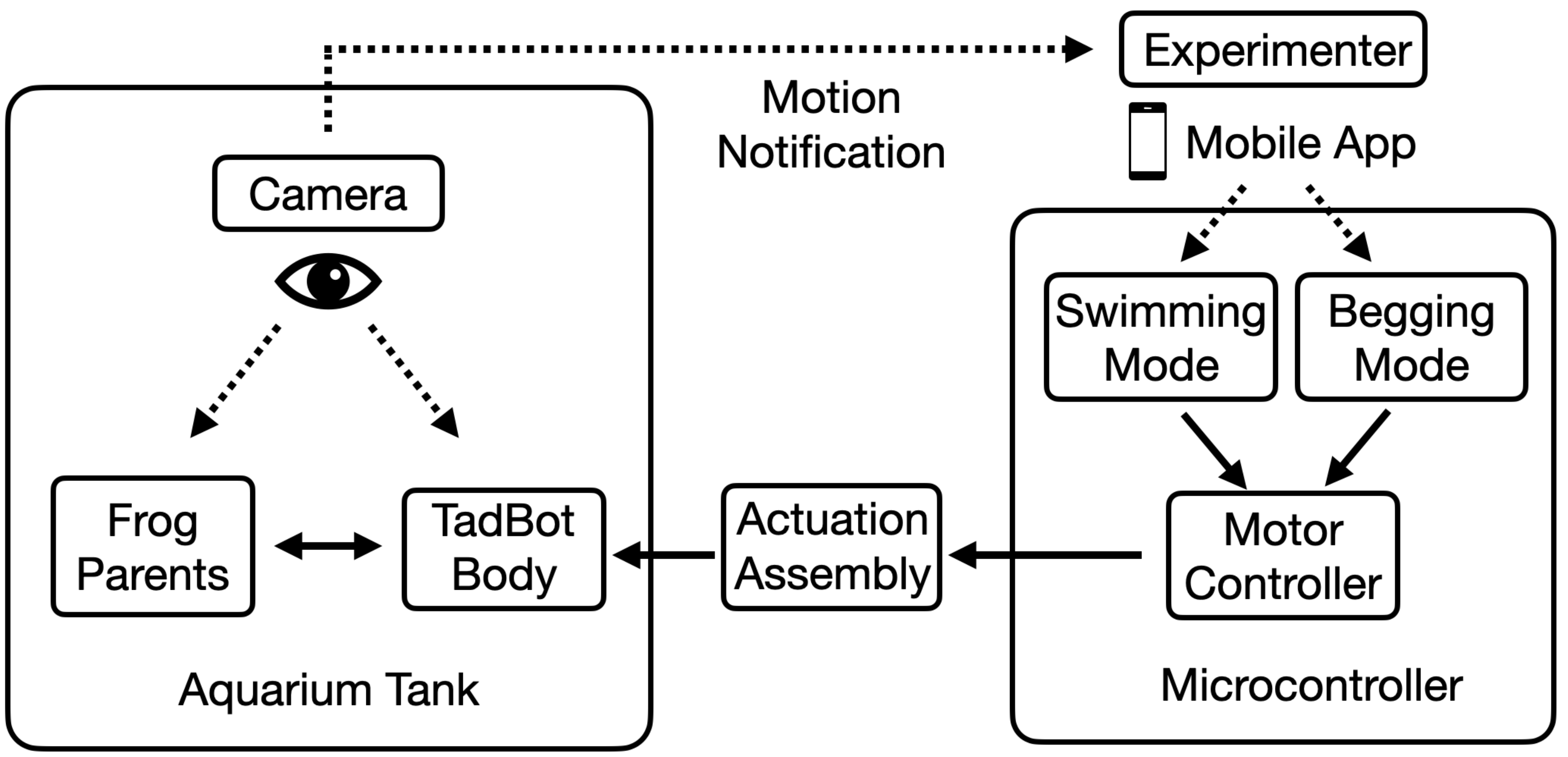}
\caption{System diagram of the experiment. A camera observes the interaction between the frog parents and the TadBot. Once a motion is detected, a notification is sent to the experimenter, who can control the operation of TadBot through a mobile app.}\label{exp}
\vspace{-6pt}
\end{figure}

Wyze v3 cameras were adhered by velcro onto the side of the Exoterra tanks and suspended above the tadpole canisters, with the face of the camera approximately \SI{17.5}{\centi\meter} above the bottom of the canister. Cameras were given 256 GB SD cards to store a month of recording. The camera observation methods are described in previous work \cite{goolsby2023home}.

A motion detection notification is sent to the experimenters when the Wyze camera detects any frog movements. Then, based on the reaction of the frogs, and at the experimenter's discretion, TadBot is activated using one of two modes through the use of a mobile app. The two modes are intended to make TadBot more analogous to living tadpoles, with different paradigms of movement to reflect affiliative and neutral behaviors.
The \emph{swimming} mode commands TadBot to intermittently wiggle its tail in 8\,Hz (15 seconds on, 10 seconds off, repeat 3 times). The \emph{begging} mode issues a wiggling signal at \SI{16}{Hz} with the same pattern. Using swimming mode versus begging mode enables the experimenter to test which frequencies a frog uses to make care decisions. 
The microcontrollers used in this experiment, Particle Argons, are connected to the cloud through local WiFi (Fig.~\ref{exp}).

To determine the influence of begging on parental decision-making, poison frog families are placed into randomized trials after a biological tadpole has been deposited into a nursery and is confirmed to be fed at least once. Parents are exposed to, in randomized order: a cross-foster living tadpole (a positive control); a TadBot with no actuation assembly (negative control); and a TadBot with an actuation assembly (experimental group). Parents are provided with two weeks per experimental stimulus, the approximate time necessary to observe repeated bouts of paternal monitoring, and at least one bout of maternal provisioning. 

\vspace{-2mm}
\section{Conclusions and Future Work}
\label{conclusions}
We have described how a tadpole-mimetic robot, TadBot, was developed for studying the parenting behaviors of \PF poison frogs. TadBots physically resemble \PF tadpoles, operate in tadpole nurseries, and mimic the tadpoles' begging behavior which includes vigorous tail wiggling at characteristic frequencies. To evaluate parenting response to begging intensity, an experiment involving multiple \PF parents has been constructed in which TadBots are substituted for live tadpoles and controlled remotely using cameras and a mobile app to produce swimming or begging motions.

A preliminary study is in progress with $n=4$ parenting pairs. In all of these pairs we have observed on multiple occasions that fathers, after observing begging signals from TadBots, have begun to coordinate care, distinguished by calling and soliciting mothers to tadpole nurseries to provision them (Fig.~\ref{fig1}C). The calls are consistent with those documented in \cite{moss2023evolution}. Examples of the behaviors can be seen in videos posted at \small{\url{http://bdml.stanford.edu/TadBot}} to accompany this paper. 
Tadpoles beg to both their mothers and fathers \cite{yoshioka2016evidence}.
Mothers decide to provision eggs based on signals that are not entirely clear but may include vibrational signals from the tadpoles (which may include physical contact) and acoustic signals from the fathers. Thus far we have observed the mothers visiting the begging tadbot, but no unfertilized eggs were deposited
More longitudinal work is necessary to determine quantitatively the amount of care that robotic tadpoles receive versus  biological offspring. 

Ongoing work includes refinement of TadBots based on the preliminary study. In the next generation we will employ a softer and more flexible tubing for the tendon, to allow more movement inside of canister -- in part so that TadBots can more convincingly vibrate their heads against the mothers. Another refinement will be to coat the skin with a hydrogel, as described in \cite{kight2023decoupling}, to increase tactile realism.

\subsubsection{Acknowledgements.} The authors acknowledge that this research was conducted on the ancestral lands of the Muwekma Ohlone people at Stanford. We thank the Laboratory of Organismal Biology for support, assistance, and input. We thank Dave Ramirez and Madison Lacey for their continued care of our poison
frog colony.
\vspace{-2mm}
\paragraph{Funding:} T.G.C was supported by a NSF Graduate Research Fellowship. B.C.G. was supported by an HHMI Gilliam Fellowship (GT15685) and a NIH Cellular Molecular Biology Training Grant (T32GM007276). This research was funded with grants 
from the NIH (DP2HD102042) and the New York Stem Cell Foundation. LAO is a New York Stem Cell Foundation–Robertson Investigator.



%
%
%
\vspace{-3mm}
\bibliographystyle{splncs04}
\bibliography{TadBot.bib}

\begin{thebibliography}{10}
\providecommand{\url}[1]{\texttt{#1}}
\providecommand{\urlprefix}{URL }
\providecommand{\doi}[1]{https://doi.org/#1}

\bibitem{brown2010key}
Brown, J.L., Morales, V., Summers, K.: A key ecological trait drove the
  evolution of biparental care and monogamy in an amphibian. The american
  naturalist  \textbf{175}(4),  436--446 (2010)

\bibitem{brown2011taxonomic}
Brown, J.L., Twomey, E., Amezquita, A., De~Souza, M.B., Caldwell, J.P.,
  Loetters, S., Von~May, R., Melo-Sampaio, P.R., Mejia-Vargas, D., Perez-Pena,
  P., et~al.: A taxonomic revision of the neotropical poison frog genus
  ranitomeya (amphibia: Dendrobatidae). Zootaxa  \textbf{3083}(1),  1--120
  (2011)

\bibitem{brown2008divergence}
Brown, J., Morales, V., Summers, K.: Divergence in parental care, habitat
  selection and larval life history between two species of peruvian poison
  frogs: an experimental analysis. Journal of Evolutionary Biology
  \textbf{21}(6),  1534--1543 (2008)

\bibitem{caldwell2010vibrational}
Caldwell, M.S., Johnston, G.R., McDaniel, J.G., Warkentin, K.M.: Vibrational
  signaling in the agonistic interactions of red-eyed treefrogs. Current
  Biology  \textbf{20}(11),  1012--1017 (2010)

\bibitem{coss2022can}
Coss, D.A., Ryan, M.J., Page, R.A., Hunter, K.L., Taylor, R.C.: Can you
  hear/see me? multisensory integration of signals does not always facilitate
  mate choice. Behavioral Ecology  \textbf{33}(5),  903--911 (2022)

\bibitem{dugas2017tadpole}
Dugas, M.B., Strickler, S., Stynoski, J.L.: Tadpole begging reveals high
  quality. Journal of evolutionary biology  \textbf{30}(5),  1024--1033 (2017)

\bibitem{goolsby2023home}
Goolsby, B.C., Fischer, M.T., Pareja-Mejia, D., Lewis, A.R., Raboisson, G.,
  O'Connell, L.A.: Home security cameras as a tool for behavior observations
  and science equity. bioRxiv pp. 2023--04 (2023)

\bibitem{gosner1960simplified}
Gosner, K.L.: A simplified table for staging anuran embryos and larvae with
  notes on identification. Herpetologica  \textbf{16}(3),  183--190 (1960)

\bibitem{hill2022biotremology}
Hill, P.S., Mazzoni, V., Stritih-Peljhan, N., Virant-Doberlet, M., Wessel, A.:
  Biotremology: physiology, ecology, and evolution. Springer (2022)

\bibitem{jung2019red}
Jung, J., Kim, S.J., P{\'e}rez~Arias, S.M., McDaniel, J.G., Warkentin, K.M.:
  How do red-eyed treefrog embryos sense motion in predator attacks? assessing
  the role of vestibular mechanoreception. Journal of Experimental Biology
  \textbf{222}(21),  jeb206052 (2019)

\bibitem{jungfer1996reproduction}
Jungfer, K.H.: Reproduction and parental care of the coronated treefrog,
  anotheca spinosa (steindachner, 1864)(anura: Hylidae). Herpetologica pp.
  25--32 (1996)

\bibitem{kam2002female}
Kam, Y.C., Yang, H.W.: Female--offspring communication in a taiwanese tree
  frog, chirixalus eiffingeri (anura: Rhacophoridae). Animal Behaviour
  \textbf{64}(6),  881--886 (2002)

\bibitem{kight2023decoupling}
Kight, A., Pirozzi, I., Liang, X., McElhinney, D.B., Han, A.K., Dual, S.A.,
  Cutkosky, M.: Decoupling transmission and transduction for improved
  durability of highly stretchable, soft strain sensing: Applications in human
  health monitoring. Sensors  \textbf{23}(4), ~1955 (2023)

\bibitem{klein2012robots}
Klein, B.A., Stein, J., Taylor, R.C.: Robots in the service of animal behavior.
  Communicative \& integrative biology  \textbf{5}(5),  466--472 (2012)

\bibitem{moss2023evolution}
Moss, J.B., Tumulty, J.P., Fischer, E.K.: Evolution of acoustic signals
  associated with cooperative parental behavior in a poison frog. Proceedings
  of the National Academy of Sciences  \textbf{120}(17),  e2218956120 (2023)

\bibitem{ringler2017adopt}
Ringler, E., Barbara~Beck, K., Weinlein, S., Huber, L., Ringler, M.: Adopt,
  ignore, or kill? male poison frogs adjust parental decisions according to
  their territorial status. Scientific Reports  \textbf{7}(1), ~1--6 (2017)

\bibitem{ringler2016sex}
Ringler, E., Pa{\v{s}}ukonis, A., Ringler, M., Huber, L.: Sex-specific
  offspring discrimination reflects respective risks and costs of misdirected
  care in a poison frog. Animal Behaviour  \textbf{114},  173--179 (2016)

\bibitem{serrano2021tadpole}
Serrano-Rojas, S.J., Pa{\v{s}}ukonis, A.: Tadpole-transporting frogs use
  stagnant water odor to find pools in the rainforest. Journal of Experimental
  Biology  \textbf{224}(21),  jeb243122 (2021)

\bibitem{stynoski2009discrimination}
Stynoski, J.L.: Discrimination of offspring by indirect recognition in an
  egg-feeding dendrobatid frog, oophaga pumilio. Animal Behaviour
  \textbf{78}(6),  1351--1356 (2009)

\bibitem{taylor2008faux}
Taylor, R.C., Klein, B.A., Stein, J., Ryan, M.J.: Faux frogs: multimodal
  signalling and the value of robotics in animal behaviour. Animal Behaviour
  (2008)

\bibitem{wright2007evolution}
Wright, J., Leonard, M.L.: The evolution of begging: competition, cooperation
  and communication. Springer Science \& Business Media (2007)

\bibitem{yoshioka2016evidence}
Yoshioka, M., Meeks, C., Summers, K.: Evidence for begging as an honest signal
  of offspring need in the biparental mimic poison frog. Animal Behaviour
  \textbf{113},  1--11 (2016)

\end{thebibliography}

\end{document}